\documentclass[11pt,a4paper]{article}
\usepackage[hyperref]{naaclhlt2019}
\usepackage{times}  
\usepackage{helvet}  
\usepackage{courier}  
\usepackage{url}  
\usepackage{graphicx}  
\usepackage{paralist}
\usepackage{subfig}
\usepackage{amsmath}
\usepackage{latexsym}
\usepackage{booktabs}
\usepackage{tabularx}
\usepackage{multirow}
\usepackage{verbatim}
\usepackage{xcolor}
\aclfinalcopy 

\title{Improving Neural Machine Translation with \\ Pre-trained Representation}
\author{Rongxiang Weng$^{1}$, Heng Yu$^{1}$, Shujian Huang$^{2}$, Weihua Luo$^{1}$, and Jiajun Chen$^{2}$ \\
$^1$Machine Intelligence Technology Lab, Alibaba DAMO Academy \\
$^2$National Key Laboratory for Novel Software Technology, Nanjing University \\
{\tt wengrongxiang@gmail.com, \{yuheng.yh,weihua.luowh\}@alibaba-inc.com, }\\
{\tt\{huangsj,chenjj\}@nju.edu.cn}}

\date{}
\hypersetup{draft}
\begin{document}
\maketitle
\begin{abstract}
Monolingual data has been demonstrated to be helpful in improving the translation quality of neural machine translation (NMT).
The current methods stay at the usage of word-level knowledge, such as generating synthetic parallel data or extracting information from word embedding. In contrast, the power of sentence-level contextual knowledge which is more complex and diverse, playing an important role in natural language generation, has not been fully exploited.
In this paper, we propose a novel structure which could leverage monolingual data to acquire sentence-level contextual representations. 
Then, we design a framework for integrating both source and target sentence-level representations into NMT model to improve the translation quality.
Experimental results on Chinese-English, German-English machine translation tasks show that our proposed model achieves improvement over strong Transformer baselines, while experiments on English-Turkish further demonstrate the effectiveness of our approach in the low-resource scenario. \footnote{In Progress}
\end{abstract}

\section{Introduction}
Neural machine translation (NMT) based on an encoder-decoder framework~\cite{sutskever2014sequence,Cho2014Learning,Bahdanau2015Neural,Luong2015Effective} has obtained state-of-the-art performances on many language pairs~\cite{deng2018alibaba}.
Various advanced neural architectures have been explored for NMT under this framework, such as recurrent neural network (RNN)~\cite[RNNSearch]{Bahdanau2015Neural,Luong2015Effective}, convolutional neural network (CNN)~\cite[Conv-S2S]{gehring2016convolutional}, self-attention network~\cite[Transformer]{vaswani2017attention}.

Currently, most NMT systems only utilize the sentence-aligned parallel corpus for model training.
Monolingual data, which is larger and easier to collect, is not fully utilized limiting the capacity of the NMT model.
Recently, several successful attempts have been made to improve NMT by incorporating monolingual data~\cite{gulcehre2015using,sennrich2016improving,Zhang2016Exploiting,poncelas2018investigating}, and reported promising improvements.

However, these studies only focus on the usage of word-level information, e.g. extracting information from word embedding. Meanwhile, several methods such as ELMo \cite{peters2018deep}, GPT \cite{radford2018improving} and BERT \cite{devlin2018bert} have gained tremendous success by modeling sentence representation. Thus, the upcoming question is whether the sentence representation is useful in machine translation, and these methods of acquiring and integrating could be applied directly to it.

In this paper, we compare these methods and demonstrate it is non-trivial to propose new methods to acquire sentence representation and integrate it into NMT for three main reasons:
First, due to the limited amount of parallel data, compared with word level representation, the NMT models may not generate appropriate sentence representation, which is more complex and diverse. However, sentence representation is important in text generation. So, the pre-trained sentence representation could be an excellent complement to provide proper information for NMT.
Second, the current methods of acquiring pre-trained representations don't fully model the sequential information in the sentence, which is a fundamental factor for modeling sentence.
Last, machine translation, which is a \textit{bilingual language generation} task, has not been well studied for how to integrate external knowledge into it. For example, the generation process is on the fly during the inference stage, most current methods could not benefit the decoder with the partially translated inputs.

To address the challenges above, we propose a solution to better leverage the pre-trained sentence representation to improve NMT as follows:
First, we propose a \textit{bi-directional self-attention language model} (BSLM) to acquire sentence representation which is trained by the monolingual data. 
The BSLM could capture the forward and backward sequential information from a sentence to build better representations.

Then, we design a framework to integrate effectively task-specific information from pre-trained representation into NMT on both source and target sides. We propose a \textit{weighted-fusion mechanism} for fusing the task-specific representation into the encoder. For the decoder, we use a \textit{knowledge transfer paradigm} to learn task-specific knowledge from the pre-trained representation.
This framework could be applied to various neural structures based on the encoder-decoder framework~\cite{Bahdanau2015Neural,gehring2016convolutional,vaswani2017attention} and language generation tasks.

To demonstrate the effectiveness of our approach, we implement the proposed approach on the current state-of-the-art Transformer model~\cite{vaswani2017attention}. Experimental results on Chinese-English and German-English translation tasks show that our approach improves over the Transformer baseline on the standard data-sets. 
Moreover, we show that on low resource language pair like English-Turkish, our approach leads to more improvements. 

\section{Approach}
In this section, we will introduce our proposed solution for acquiring and integrating sentence representation in detail. 
First, we propose a \textit{bi-directional self-attention language model} (BSLM) trained on large scale unlabeled monolingual sentences to get \textit{sentence representation}. 
Next, we introduce two individual methods to employ the generated representation: a \textit{weighted-fusion mechanism} and a \textit{knowledge transfer paradigm} to enhance the encoder and decoder by fusing and learning the information from the sentence representation, respectively.
\begin{figure}
    \centering
    \includegraphics[scale = 0.45]{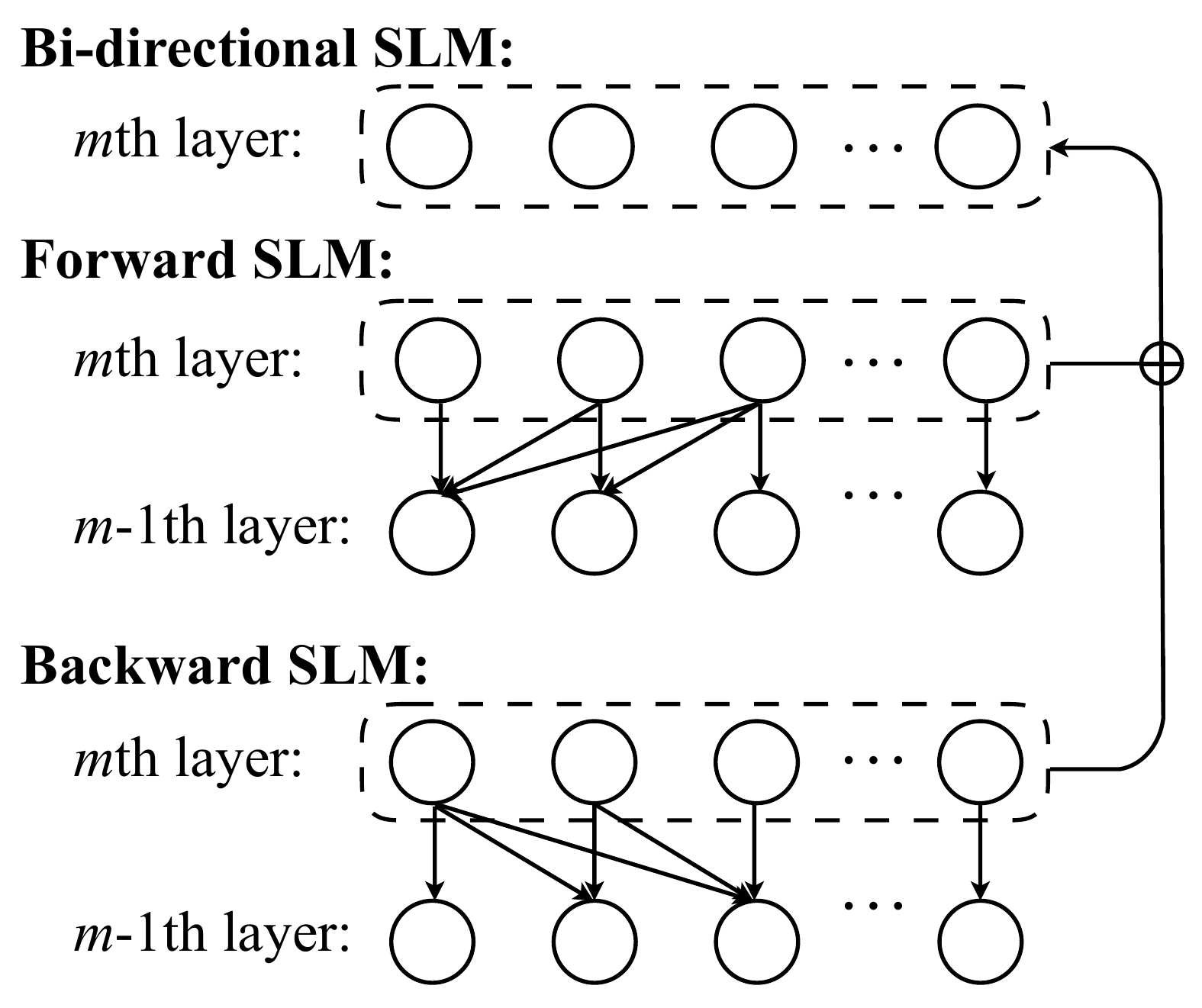}
    \caption{Overview of the bidirectional self-attention language model (BSLM).}
    \label{fig:slm}
\end{figure}

\subsection{Bi-directional Self-Attention Language Model}
\label{bslm}
Previous studies~\cite{peters2018deep,radford2018improving,devlin2018bert} indicate that sentence representation generated by the pre-trained model could help the downstream tasks.
Furthermore, compared with traditional RNN~\cite{jozefowicz2016exploring,melis2017state,peters2018deep} structure, self-attention network (SAN), which could achieve wider and deeper information, has a stronger ability to extract features~\cite{tang2018self}.

However, most of these SAN based pre-trained methods ignore the sequential information which is a fundamental factor for sentence modeling.
To fully acquire sentence representation with sequential information, we propose a bi-directional self-attention language model (BSLM).
Our BSLM model consists of a forward and a backward self-attention language model~(SLM) which can capture the forward and backward sequential information by using the directional matrix~\cite{shen2017disan} augments sequential relation~(see Figure \ref{fig:slm}). 
Moreover, inspired by~\newcite{wang2018multi} and \newcite{dou2018exploiting}, we gather all layers' representations from BSLM which contain different aspects of information for a sentence.

Specifically, the forward SLM has $M$ layers and the representation in the $m$th layer is defined as:
\begin{align}
\overrightarrow{\textbf{R}}^{L}_{m} = (\overrightarrow{\textbf{r}}^{L}_{m,1},\overrightarrow{\textbf{r}}^{L}_{m,2}, \cdots,\overrightarrow{\textbf{r}}^{L}_{m,k}, \cdots,\overrightarrow{\textbf{r}}^{L}_{m,K} ), \nonumber
\end{align}
where $K$ is the number of vectors in the $\overrightarrow{\textbf{R}}^{L}_{m}$. The representation matrix is computed by:
\begin{align}
\overrightarrow{\textbf{R}}^{L}_{m}&=\text{LN}(\text{FFN}(\overrightarrow{\textbf{H}}_{m})+\overrightarrow{\textbf{R}}^{L}_{m-1}),  \label{eq: layernorm}
\end{align}

The $\overrightarrow{\textbf{H}}_{m}$ is calculated by:
\begin{align}
\overrightarrow{\textbf{H}}_{m}&=\text{MultiHead}(\textbf{Q},\textbf{K},\textbf{V}) \nonumber \\ 
&=\text{Concat}(\textbf{head}_1,\cdots,\textbf{head}_H), \label{eq: multi-head}
\end{align}
where $\textbf{Q}$, $\textbf{K}$, $\textbf{V}$ are query, key and value matrixes that are equal to $\overrightarrow{\textbf{R}}^{L}_{m-1}$.

We introduce a \textit{directional matrix} proposed by~\newcite{shen2017disan} in vanilla self-attention network~\cite{vaswani2017attention}, in which a mask matrix is used to get the sequential information by covering the rear words for each input word. So a single attention head is:
\begin{align}
\textbf{head}_\text{h}&=\text{DirAtt}(\textbf{Q},\textbf{K},\textbf{V})  \nonumber \\
&=\text{softmax}(\frac{\textbf{Q}\textbf{K}^\top}{\sqrt{d_k}}+\textbf{Mask})\textbf{V}. \label{eq: diratt}
\end{align}
The $\textbf{Mask}$ is calculated by:
\begin{align}
\textbf{Mask}_{c,v}=\left\{
\begin{aligned}
-\infty&, \ c<v\\
0&, \ otherwise,
\end{aligned}
\right.
\end{align}
the row and column of the $\textbf{Mask}$ are equal to $K$.

Typically, $\overrightarrow{\textbf{r}}^{L}_{1,k}$ from $\overrightarrow{\textbf{R}}^{L}_{1}$ is defined as $\overrightarrow{\textbf{r}}^{L}_{1,k} = \text{emb}^{L}(w_{k})$, where $w_{k}$ is a input word and $\text{emb}^{L}(w_{k})$ is the word embedding of $w_{k}$.
Therefore, we have $\overrightarrow{\textbf{R}}^{L}_{1}$ with the given input: $\textbf{w}= (w_{1},w_{2}, \cdots,w_{k}, \cdots,w_{K} ).$

In the training stage, the $k$th output word is computed by maximizing the conditional probability, which is defined as: 
\begin{align}
P(w_{k}|w_{<k})&=\text{softmax}(\overrightarrow{\textbf{r}}_{M,k}), \label{eq: prob}
\end{align}
The SLM is optimized by maximizing the likelihood, defined as:
\begin{equation}
    \mathcal{L}_{\text{L}}=\frac{1}{K}\sum_{k=1}^{K}\log P(w_{k}|w_{<k}).
\end{equation}
Then, the representation from the forward SLM is:
\begin{align}
\overrightarrow{\textbf{R}}^{L}=(\overrightarrow{\textbf{R}}^{L}_{1},\cdots,\overrightarrow{\textbf{R}}^{L}_{m},\cdots,\overrightarrow{\textbf{R}}^{L}_{M}). \nonumber
\end{align}

The structure and training process of the backward SLM are similar to the forward one. At last, we can get the sentence representation $\textbf{R}^{L}$ containing forward and backward sequential information by 
adding the $\overrightarrow{\textbf{R}}^{L}$ and $\overleftarrow{\textbf{R}}^{L}$, which are from the forward and backward SLMs, respectively. 

\begin{figure*}[ht]
    \centering
    \subfloat[The weighted-fusion mechanism used in the encoder.]{
     \label{fig:Left2right}
    \includegraphics[scale=.45]{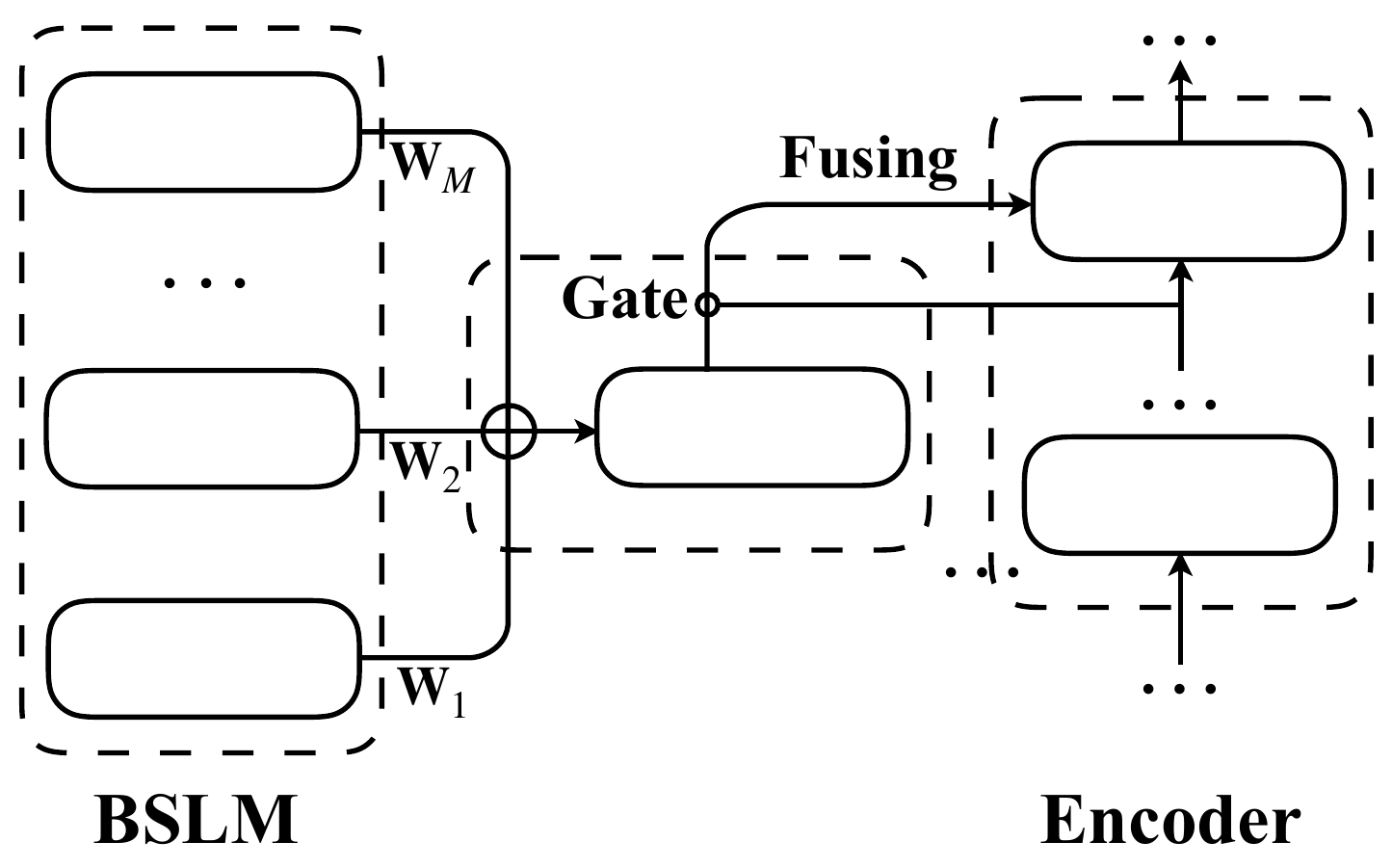}
    } ~~~~~~~
    \subfloat[The knoledge transfer paradigm used in the decoder.]{
     \label{fig:bidirect}
    \includegraphics[scale=.45]{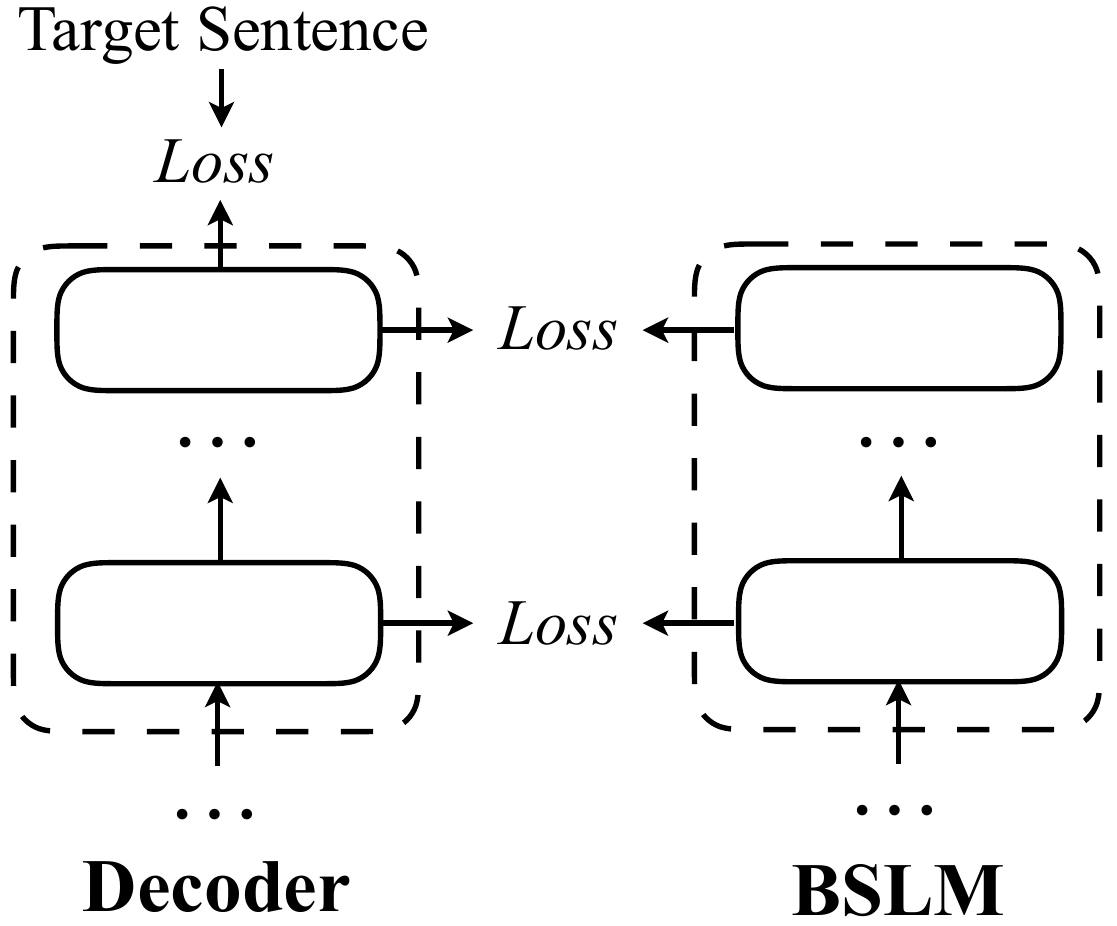}}
    \label{fig:intro_example}
    \caption{Overview of the proposed integration framework.}
\end{figure*}
\subsection{Integrating Sentence Representation in Transformer}
Due to the limited amount of parallel data, it is hard for the Transformer to generate appropriate sentence representation. The pre-trained BSLM could be an excellent complement to provide Transformer with proper sentence level information.

However, the previous integration methods like initializing parameters from pre-trained model may not suit for the machine translation which is a bilingual generation task.
The general representation is quite different from the task-specific representation of NMT model.
Thus, we propose a novel integration framework to fully utilize pre-trained sentence representation in the NMT model.

\paragraph{Weighted-fusion Mechanism.} 
On the source side, partially inspired by \newcite{peters2018deep}, we propose to use a \textit{weighted-fusion mechanism} to generate \textit{task-specific representation} from the pre-trained sentence representations, and fuse them into the encoder of Transformer through a gating network.

Different from previous works~\cite{ramachandran2016unsupervised,peters2018deep,radford2018improving}, the proposed feature-based method can make a \textit{deep fusion} which could incorporate appropriate information into each layer of the encoder, that is, Transformer can access the specific surface information in lower layers and the latent representation for its higher layers. 


Formally, for the $n$th layer of the encoder, the vanilla representation $\textbf{R}^{S}_n$ is computed by Equation \ref{eq: layernorm}-\ref{eq: diratt}, in which the $\textbf{Mask}$ is a zero matrix. 
The sentence representation $\textbf{R}^{L}$ from BSLM will be fused into $\textbf{R}^{S}_n$ by:
\begin{align}
\textbf{R}^{S}_{n} &= \textbf{R}^{S}_{n}+\theta_{n} * \textbf{R}^{W}_{n}, \end{align}
$\theta_{n}$ is a dynamic gate which means how much the current layer needs the extra information. The gate is computed by each state from the $\textbf{R}^{S}_n$:
\begin{align}
\theta_{n} &=\text{sigmoid}(\overline{\textbf{r}}^{S}_{n}), \\
\overline{\textbf{r}}^{S}_{n} &=\frac{1}{I}\sum^{I}_{i=1}\textbf{r}^{S}_{n,i},
\end{align}
$I$ is the length of the source sentence $\textbf{x}$. We denote $\textbf{r}^{S}_{n,i}$ to be the word embedding of the corresponding word $x_i$ in the $\textbf{x}$.
The task-specific representation $\textbf{R}^{W}_{n}$ is calculated by:
\begin{align}
\textbf{R}^{W}_{n} &=\sum^{M}_{m=1}(\textbf{W}_{n, m}*\textbf{R}^{L}_{m}). \label{eq: weight}
\end{align}
$\textbf{W}_{n, m}$ is a trainable weight representing the importance of $\textbf{R}^{L}_{m}$ learned in the training process.

\paragraph{Knowledge Transfer Paradigm.} 
Compared with the encoder, the decoder is difficult to exploit the pre-trained knowledge whenever using feature-based or fune-tuning methods~\cite{devlin2018bert,peters2018deep}.

For example, the \textit{exposure bias}~\cite{lee2018deterministic,wu2018beyond} leads to that the ground-truth sentence representation which is generated by the reference is not available in the inference stage. 
On the other hand, the decoding process involves reading the source side representations, these parameters are hard to initialized by pre-trained models. It leads to the fine-tuning methods do not work well here~\cite{radford2018improving}.

Therefore, we propose a simple \textit{knowledge transfer paradigm} to capture the bi-directional sequential information by learning the pre-trained sentence representation in the training stage. This method could transfer the knowledge from monolingual data to the NMT model and could avoid the problems mentioned above.

We design an auxiliary learning objective besides traditional translation objective, which is meant to transfer bi-directional sentence knowledge from BSLM to the NMT model. Formally, the auxiliary knowledge transfer objective of our proposed paradigm is:

\begin{align}
\mathcal{L}_{\text{E}}=\frac{1}{J}\sum^{N}_{n=1}\sum^{J}_{j=1}||\textbf{r}^{T}_{n,j}-\textbf{r}^{L}_{n,j}||^{2}_{2}, \label{eq: multi-task}
\end{align}
where the $J$ is the length of given target sentence $\textbf{y}$, the $N$ is number of layers. The $\textbf{r}^{T}_{n,j}$ and $\textbf{r}^{L}_{n,j}$ are from the decoder and BSLM, respectively. Note that the number of layers of BSLM and the decoder should be the same to utilize the knowledge transfer method.


The translation model is optimized by maximizing the likelihood of the  $\textbf{y}$ given source sentence $\textbf{x}$:
\begin{align}
\mathcal{L}_{\text{M}}=\frac{1}{J}\sum_{j=1}^{J}\log P(y_{j}|y_{<j},\textbf{x}). \label{eq: prob2}
\end{align}
The $P(y_{j}|y_{<j},\textbf{x})$ is computed by Equation \ref{eq: prob}.
Finally, the loss function of our model is:
\begin{align}
\mathcal{L}_{\text{T}}=\mathcal{L}_{\text{M}}+\mathcal{L}_{\text{E}}. \label{eq: joint}
\end{align}

\section{Experiment}

\begin{table*}[ht]
\centering
\begin{tabular}{c|l||cccc|c|c}
\hline
\#&\textbf{Model} &\textbf{MT02}&\textbf{MT03} & \textbf{MT04} & \textbf{MT05}&\textbf{Average} &$\Delta$ \\
\hline
1&RNNSearch~\cite{Luong2015Effective} &	N/A&28.38  &30.85 &  26.78&$-$&$-$ \\
\hline
2&\cite{sennrich2016improving} &36.95 &36.80  &37.99  &35.33 &$-$ &$-$ \\
3&\cite{Zhang2016Exploiting} &N/A &33.38 &34.30 &31.57 & $-$&$-$ \\
4&\cite{cheng2016semi} &38.78&38.32&38.49
 &36.45 &$-$ &$-$ \\
5&\cite{zhang2018joint} &N/A&43.26&N/A
 &41.61 &$-$ &$-$ \\
\hline
\hline
6&Transformer &44.77	&44.93 & 45.81 &43.04  &44.59 &$-$ \\
\hline
7&\ \ + ELMo~\cite{peters2018deep} &45.23 &45.60 &46.26 &43.61  &45.16	&+0.57 \\

8&\ \ + GPT~\cite{radford2018improving} &44.89 &45.22 &45.99 &43.31  &44.84	&+0.25\\

9&\ \ + BERT~\cite{devlin2018bert} &45.02 &45.53 & 46.02& 43.52&45.02 &+0.43 \\
\hline
\multicolumn{8}{c}{\textit{Effectiveness of weighted-fusion mechanism used in the different layers}} \\
\hline
10&\ \ + Weighted-fusion (\textit{shallow}) &44.97 &45.21 &46.19 &43.23  &44.88 &+0.29\\
11&\ \ + Weighted-fusion (\textit{deep}) &45.46 &45.62 &46.57 &43.82  &45.34 &+0.75\\
\hline
\multicolumn{8}{c}{\textit{Effectiveness of knowledge transfer paradigm used in the different layers}} \\
\hline
12&\ \ + Knowledge Transfer (\textit{shallow})  &45.61 &45.63 &46.54 &43.86  &45.34 &+0.75 \\
13&\ \ + Knowledge Transfer (\textit{deep}) &45.71 &45.78 &46.62 &43.94  &45.45 &+0.85 \\
\hline
\multicolumn{8}{c}{\textit{Our proposed model}} \\
\hline
14&\ \ + Our Approach &\textbf{45.82} &\textbf{45.86} &\textbf{46.83} &\textbf{44.13}  &\textbf{45.61} & \textbf{+1.01}\\
\hline
\end{tabular}
\caption{Translation qualities on the ZH$\rightarrow$EN experiments. \textit{deep} and \textit{shallow} mean employing proposed methods on the all layers or the first layer, respectively. }
\label{tab:zh-en_r}
\end{table*}

\subsection{Implementation Detail}
\paragraph{Data-sets.} We conduct experiments on Chinese$\rightarrow$English (ZH$\rightarrow$EN), German$\rightarrow$English (DE$\rightarrow$EN) and English$\rightarrow$Turkish (EN$\rightarrow$TR) translation tasks. 

On the ZH$\rightarrow$EN tasks, training set consists of about 1 million sentence pairs from LDC.\footnote{include NIST2002E18, NIST2003E14, NIST2004T08, NIST2005T06}
We use \texttt{MT02} as our validation set, and  {\texttt{MT03}}, {\texttt{MT04}} and {\texttt{MT05}}  as our test sets. 
We use 8 million monolingual sentences from LDC\footnote{include NIST2003E07, NIST2004E12, NIST2005T10, NIST2006E26, NIST2007T09}.
On the DE$\rightarrow$EN tasks, We use WMT-16 as our training set, which consists of about 4.5 million sentence pairs. We use {\texttt{newstest2015}} (NST15) as our validation set, and \texttt{newstest2016} (NST16) as test sets \footnote{http://www.statmt.org/wmt17/translation-task.html}.
We use 40 million monolingual sentences from WMT-16 Common Crawl data-set.
On the EN$\rightarrow$TR tasks, We use WMT-16 as our training set, which consists of about 0.2 million sentence pairs. 
We use {\texttt{newsdev2016}} (NSD16) as our validation set and \texttt{newstest2016} (NST16) as test sets. 
We use 5 million monolingual sentences from WMT-16 Common Crawl data-set.

\paragraph{Setting.} We apply byte pair encoding (BPE)~\cite{sennrich2015neural} to encode all sentences from the parallel and monolingual data-sets and limit the vocabulary size to 32K.
Out-of-vocabulary words are denoted to the special token \emph{UNK}. 

we set the dimension of input and output of all layers for the bi-directional self-attention language model (BSLM) and Transformer as 512, and that of the feed-forward layer as 2048. 
We employ eight parallel attention heads. The number of layers for BSLM and each side of the Transformer is set to 6. 
Sentence pairs are batched together by approximate sentence length. Each batch has approximately 4096 tokens. 
We use Adam~\cite{kingma2014adam} optimizer to update parameters, and the learning rate was varied under a warm-up strategy with 4000 steps~\cite{vaswani2017attention}.

After training, we use the beam search for heuristic decoding, and the beam size is set as 4. 
We measure the translation quality with the IBM-BLEU score~\cite{Papineni2002bleu}. 
We implement our model upon our in-house Transformer system.

\begin{table*}[t]
\centering

\begin{tabular}{l||cccc|c|c}
\hline
\textbf{Model} &\textbf{NST15}&\textbf{NST13}&\textbf{NST14}&\textbf{NST16} &\textbf{Average}&$\Delta$ \\
\hline
Transformer &30.33 &29.14 &28.87 &35.74 &31.25&$-$ \\
\hline
\ \ + Weighted-fusion (\textit{deep})  &31.43 &29.72 &29.46 &36.48 &31.89 &+0.64 \\
\hline
\ \ + Knowledge Transfer (\textit{deep})  &31.32 &29.61 &29.73 &36.60 &31.98 &+0.73 \\
\hline
\ \ + Our Approach &\textbf{31.72 }&\textbf{30.32} &\textbf{30.29} &\textbf{36.91} &\textbf{32.51}&\textbf{+1.26}  \\
\hline
\end{tabular}
\caption{Translation qualities on the DE$\rightarrow$EN experiments.}
\label{tab:de-en_r}
\end{table*}

\subsection{Results on Standard Data-sets}
\paragraph{Chinese$\rightarrow$English.} The results on ZH$\rightarrow$EN task are shown in Table \ref{tab:zh-en_r}. We report several results from previous studies about using monolingual data in NMT. For fair comparison, we also implement ELMo~\cite{peters2018deep}, GPT~\cite{radford2018improving} and BERT~\cite{devlin2018bert} in our Transformer system. The implement details are as follows:
\begin{itemize}
    \item ELMo: \newcite{peters2018deep} proposed a contextual representation pre-trained from Bi-LSTM based language model.
    We train a three layers Bi-LSTM language model\footnote{code: https://github.com/allenai/allennlp} by using our monolingual corpus. 
    Then, the contextual representation is concatenated on the source side of the Transformer.
    \item GPT: \newcite{radford2018improving} proposed to use a pre-trained self-attention language model to initialize the downstream tasks. 
    We implement it by using a forward SLM initialize the parameters of the Transformer's source side. 
    \item BERT: \newcite{devlin2018bert} proposed to use parameters from a pre-trained bi-directional encoder optimized by special objectives to initialize the other tasks. Following them, we initialize the encoder of our translation model with the parameters from the pre-trained BERT\footnote{code: https://github.com/codertimo/BERT-pytorch} with our monolingual data.
\end{itemize}

In Table \ref{tab:zh-en_r}, we can see that the weighted-fusion mechanism in all layers contributes to 0.75 improvements (line 11), while employing knowledge transfer paradigm for the decoder of NMT can achieve 0.85 BLEU score improvement (line 13). Our proposed model, combining both methods, significantly improve 1.01 BLEU score (line 14) over a strong Transformer baseline. 
To further test our integration framework, we also try the shallow integration (\textit{shallow}) by using proposed methods only in the first layer of the NMT network (line 10 and 12). The results show that deep integration  (\textit{deep}) , which utilizes the proposed integration framework in all layers of the Transformer, can achieve a noticeable improvement compared to their shallow counterparts.

Compared with the related work on model pre-training, ELMo~\cite{peters2018deep} doesn't yield a significant improvement (+0.57) when integrated with Transformer (line 6). We conjecture that there are two possible reasons: first, the representation from the Bi-LSTM based language model can't be deeply fused with Transformer network due to the structural difference, so the relatively shallow integration leads to the performance degradation. Second, the char-CNN~\cite{peters2018deep} in their work can't capture enough information when the input is subword~\cite{sennrich2015neural}, which limits the performance of ELMo. 

The GPT~\cite{radford2018improving} and BERT~\cite{devlin2018bert} methods improve 0.25 and 0.43 BLEU score, respectively. Compared with the results in other monolingual tasks, the improvement is relatively small. The main reason is that these methods for parameter initialization can't fully exploit the pre-trained model and may not suit for the bilingual task as machine translation.   

\paragraph{German$\rightarrow$English.} 
To show the generalization ability of our approach, we carry out experiments on DE$\rightarrow$EN translation task. 
The results are shown in Table \ref{tab:de-en_r}. With the weighted-fusion mechanism, the BLEU gain is 0.64, while the knowledge transfer paradigm leads to 0.73 improvements. The combination of both yields a further improvement (+1.26). It shows that our model can achieve improvements in different types of language pairs on large scale data-sets.

\subsection{Results on Low-resource Data-sets}
We report the results on the EN$\rightarrow$TR task which is a relatively low-resource language pair. Furthermore, we compare with the back-translation method~\cite{sennrich2016improving} which is predominant in the low-resource scenario.  
We translate the target language (TR) to source language (EN) by a reversely-trained NMT system, and copy the parallel data set to the same size. At last, we shuffle the training data which includes pseudo and parallel sentence pairs.
The different sizes of back-translated corpus are shown in Table \ref{tab:en-tr_size}. We translate 0.2M (\textit{small size}), 0.4M (\textit{medium size}) and 1M (\textit{big size}) as pseudo data-sets. The ratios of pseudo and parallel sentence pairs are 1:1, 2:1 and 5:1. 

The results are shown in Table \ref{tab:en-tr_r}, our method can improve the 0.71 BLEU score compared to the baseline system. Moreover, when combining with pseudo data-sets (+BT), our model can achieve further improvement. Typically, when using the large size pseudo corpus, our model can improve 1.77 BLEU score, which is a prominent gain in the low-resource scenario.

\begin{table}[ht]
\centering
\begin{tabular}{l||c|c|c}
\hline
\textbf{Size} &\textbf{Pseudo}&\textbf{Parallel} &\textbf{Ratio}\\
\hline
Small &0.2M &0.2M &1:1\\
\hline
Medium &0.4M &0.2M &2:1 \\
\hline
Big &1.0M &0.2M &5:1 \\
\hline
\end{tabular}
\caption{The different sizes of pseudo corpus.}
\label{tab:en-tr_size}
\end{table}

\begin{table}[ht]
\centering
\begin{tabular}{l||cc|c}
\hline
\textbf{Model} &\textbf{NSD16}&\textbf{NST16} &$\Delta$ \\
\hline
Transformer &15.19	&13.72  &$-$ \\
\hline
\ \ + Our Approach &15.55 &14.43 &+0.71 \\
\hline
\ \ + BT (\textit{small})&15.58 &14.64 &+0.92 \\
\ \ \ \ + Our Approach &15.84 &14.99 &+0.95 \\
\hline
\ \ + BT (\textit{medium})&15.70 &14.86 &+1.14 \\
\ \ \ \ + Our Approach &15.89 &15.01 &+1.29 \\
\hline
\ \ + BT (\textit{big})&16.04 &15.19 &+1.47 \\
\ \ \ \ + Our Approach &\textbf{16.36} &\textbf{15.49} &\textbf{+1.77} \\
\hline
\end{tabular}
\caption{Translation qualities on the EN$\rightarrow$TR experiments. (BT: back-translation; \textit{small}, \textit{medium} and \textit{big}: the different size of pseudo corpus.)}
\label{tab:en-tr_r}
\end{table}

\begin{table}[ht]
\centering
\begin{tabular}{l||c}
\hline
\textbf{Model} &\textbf{Speed}  \\
\hline
Transformer &15.06 \\
\hline
\ \ + ELMo~\cite{peters2018deep} &8.41 \\
\hline
\ \ + Our Approach &14.12 \\
\hline
\end{tabular}
\caption{The comparison of decoding efficiency (sentences/second).}
\label{tab:dec_efficiency}
\end{table}
\begin{table}[ht]
\centering
\begin{tabular}{l||c|c}
\hline
\textbf{Model} &\textbf{BLEU} &$\Delta$ \\
\hline
Transformer &44.59  &$-$ \\
\hline
\ \ + Uni-directional SLM &45.42 &+0.83 \\
\hline
\ \ + Bi-directional SLM &\textbf{45.61} &\textbf{+1.01} \\
\hline
\end{tabular}
\caption{The effectiveness of uni-directional and bi-directional self-attention language model~(SLM).}
\label{tab:uni_bi}
\end{table}
\begin{table}[t]
\centering
\begin{tabular}{l||c|c}
\hline
\textbf{Model} &\textbf{Side}&\textbf{BLEU} \\
\hline
Transformer &N/A&44.59 \\
\hline
\ \ + Weighted-fusion (\it{deep}) & \it{src.}&45.47  \\
\ \ + Weighted-fusion (\it{first}) & \it{trg.}& 44.66 \\
\ \ + Weighted-fusion (\it{deep}) & \it{trg.} & 43.93\\
\hline

\ \ + Knowledge Transfer (\it{deep}) & \it{trg.} &45.45\\
\ \ + Knowledge Transfer (\it{first}) & \it{src.} & 44.92  \\
\ \ + Knowledge Transfer (\it{deep}) & \it{src.} & 45.06  \\
\hline
\end{tabular}
\caption{The comparison of translation qualities for using the weighted-fusion and knowledge transfer methods in different sides ({\it{src.}}: source; {\it{trg.}}: target) on ZH-EN task.}
\label{tab:zh-en_source_a}
\end{table}

\subsection{Decoding Efficiency} \label{ssec:deceff}
Our BSLM is highly parallelizable which could be integrated into the Transformer without losing efficiency.
In order to verify this, we compare the decoding efficiency of the baseline, our model and ELMo~\cite{peters2018deep}.

The results are shown in Table \ref{tab:dec_efficiency}, our model leads to a very little drop in decoding efficiency compared to the baseline (15.06 sentences/second vs. 14.12 sentences/second) because the self-attention structure of our model can generate sentence representation in parallel. Whereas for ELMo, the efficiency decline drastically (8.41 sentences/second) due to the auto-regressive structure of LSTM.

\section{Analysis}
\subsection{Uni-direction Vs. Bi-direction}
We compare the effectiveness of uni-directional and bi-directional self-attention language model (SLM) used on the source side. The uni-directional SLM is the forward SLM described in Section \ref{bslm}. The results are shown in Table \ref{tab:uni_bi}, the bi-directional SLM (+1.01) outperforms the uni-directional SLM (+0.83), which verifies the combination power of forward and backward linguistic information captured by bi-directional SLM.

\subsection{Impact of Different Integration Schema}
In this section, we analyze the different integration schema for our BSLM. Specifically, we switch the integration mechanism on source and target side by employing the \textit{knowledge transfer paradigm} on the source side and the weighted-fusion mechanism on the target side. So the knowledge transfer paradigm is used to optimize the source representation directly. And on the target side, we generate the representation by feeding the \textit{partially translated part} to the BSLM and fuse them into the decoder at each time step.

The results are shown in Table \ref{tab:zh-en_source_a}. The knowledge transfer paradigm on the source side can improve the performance to some extent by refining source side representation but the impact is relatively small comparing to the weighted-fusion mechanism. On the target side, the BLEU score decreases 0.66 when using weighted-fusion in all layers. The problem is that the representation generated by the partially translated part is incomplete and may contain wrong information.
This will negatively influence the decoder, especially in higher layers. 
According to this phenomena, the category of methods which directly fuses pre-trained representation into the target side may not work well.

\subsection{Visualization of Weights}
\begin{figure}
\centering
\includegraphics[scale = 0.26]{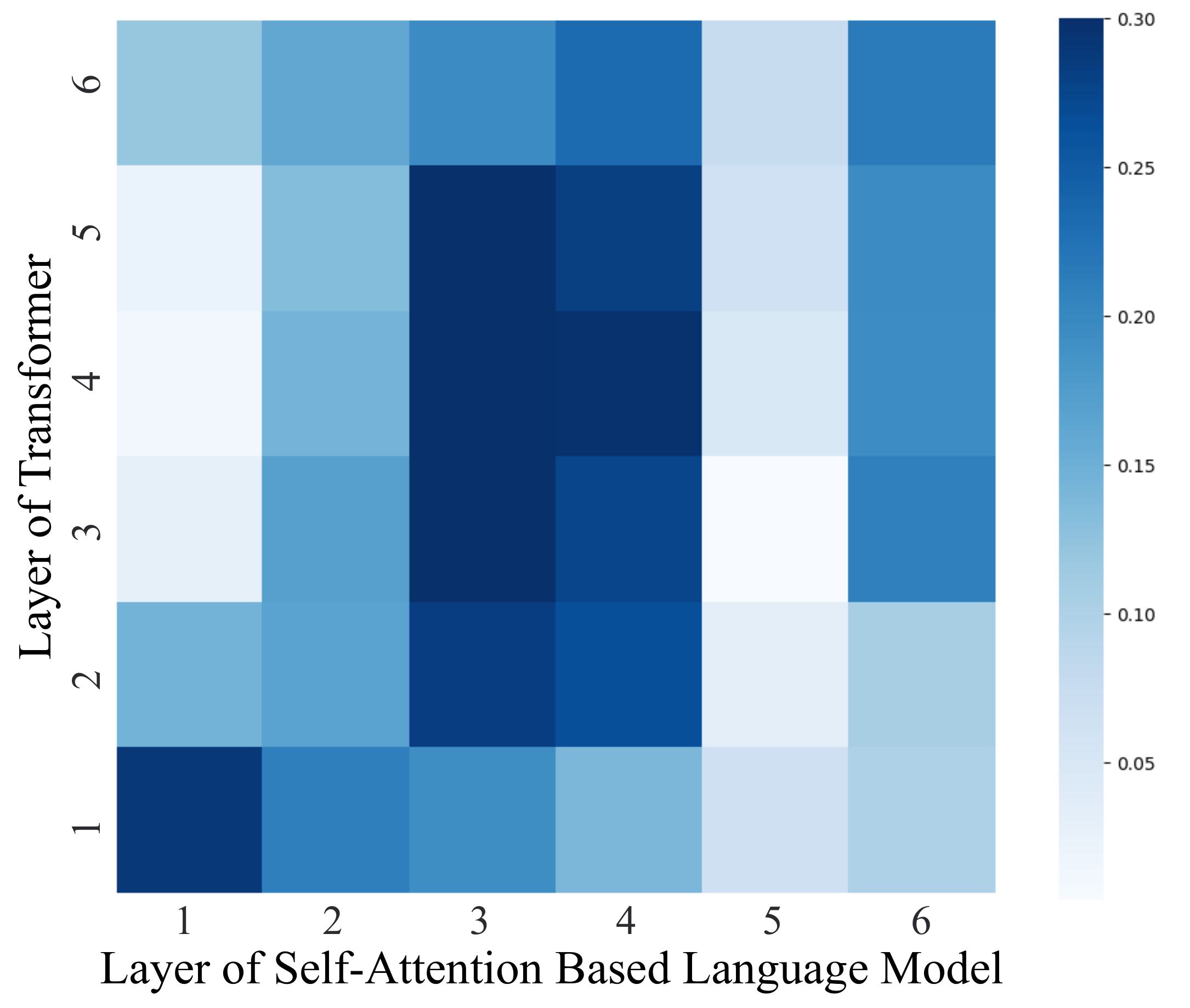}
\caption{\label{fig:heat_flow} The heat map for the learned weights of the weighted-fusion mechanism.}
\end{figure}

We draw the heat map for the learned weights of the weighted-fusion mechanism (Equation \ref{eq: weight}). 
As shown in Figure \ref{fig:heat_flow}, The column axis is the layer of Transformer and row axis is the layer of the BSLM. The number in each row is normalized. 
The map shows a good correlation between the fusion weights for different layers of Transformer decoder and BSLM. It illustrates the advantage of BSLM in the integration with Transformer from another perspective: with the similar network structure, the lower layers of BSLM can be used to model specific surface information, while higher layers provide latent representation.

\section{Related Work}

Several successful attempts have been made to utilize monolingual data in NMT directly.
\newcite{sennrich2016improving} propose to use back-translation to generate synthetic parallel data from monolingual data for NMT.
\newcite{currey2017copied} propose a copy mechanism to copy fragments of sentences from monolingual data to translated outputs directly.
\newcite{zhang2018joint} propose a joint learning of source-to-target and target-to-source NMT models, 
generating pseudo parallel data using monolingual data.
However, directly using synthetic data suffers from potential noise propagation problem. Our model only learns sentence representation rely on the gold inputs, which can avoid this problem.

On applying the pre-trained model for NMT, \newcite{di2017can} use source side pre-trained embedding and integrate it into NMT with a mix-sum/gating mechanism. They only focus on improving the source side's representation, leaving the target side information largely ignored.
\newcite{ramachandran2016unsupervised} propose to initialize the parameters of NMT by a pre-trained language model. However, these methods lack a lasting impact on training, leaving the information from pre-trained models less exploited. Our proposed solutions can generate sentence representations which can be applied on both sides of the NMT models, resulting in much better performance in terms of translation quality.


In the NLP field, sentence representation has been widely explored.
\newcite{peters2018deep} introduce Embedding learned from LSTM based Language Models (ELMo) and successfully apply it in question answering, textual
entailment, etc. Partially inspired by them, we propose to use the weighted-fusion mechanism to utilize the sentence representation.
Learning information from monolingual data by knowledge transfer is widely used in the several NLP tasks~\cite{tang2019distilling}. We are the first attempt in the neural machine translation.
\newcite{radford2018improving} use the parameters from a pre-trained self-attention structure language model to initialize the downstream tasks. \newcite{devlin2018bert} extend this idea by using the bi-directional encoder representation from Transformer (BERT) and introducing several training objectives. 
Different from them, we carefully studied the application of pre-trained model in the bilingual setting of machine translation, our proposed BSLM and integration mechanism take account of both source and target side monolingual data, and can fully explore the information from pre-trained model. 

\section{Conclusion}

In this paper, we propose a novel approach to better leveraging monolingual data for NMT by utilizing pre-trained sentence representations. 
The sentence representations are acquired through bi-directional self-attention language model (BSLM) and integrated into NMT network by a weighted-fusion mechanism and a knoledge transfer paradigm.
Experiments on various languages show that proposed model achieves prominent improvements on the standard data-sets as well as in the low-resource scenario.

\bibliography{naaclhlt2019}
\bibliographystyle{acl_natbib}


\end{document}